\documentclass[a4paper,twoside,french]{article}
\usepackage{rfiap2018}
\usepackage[utf8]{inputenc}
\usepackage[T1]{fontenc}
\usepackage[french]{babel}
\usepackage{times}
\usepackage{amsmath,amssymb,amsfonts}
\usepackage{rfiap2018}
\usepackage[utf8]{inputenc}
\usepackage[T1]{fontenc}
\usepackage[french]{babel}
\usepackage{times}
\usepackage{bm}
\usepackage{url}
\usepackage{graphicx,graphics}
\usepackage{booktabs}
\usepackage{multirow}
\usepackage[ruled,linesnumbered]{algorithm2e}
\usepackage{textcomp}
\usepackage{tabularx}
\usepackage[flushleft]{threeparttable}
\usepackage{cite}
\usepackage[table,dvipsnames]{xcolor}
\usepackage{soulutf8}
\usepackage{soul}
\usepackage{algpseudocode}

\newcommand{\midsepremove}{\aboverulesep = 0.2mm \belowrulesep = 0.2mm}
    \midsepremove
\newcommand{\midsepdefault}{\aboverulesep = 0.605mm \belowrulesep = 0.984mm}
    \midsepdefault

\begin{document}
\date{}
\title{\Large\bf Asservissement visuel 3D direct dans le domaine spectral}
\author{\begin{tabular}[t]{c@{\extracolsep{1em}}c@{\extracolsep{1em}}c@{\extracolsep{1em}}c@{\extracolsep{1em}}c}
Maxime Adjigble${}^1$  & Brahim Tamadazte${}^2$ & Cristiana de Farias${}^1$  & Rustam Stolkin${}^1$  & Naresh Marturi${}^1$\\
\end{tabular}
{} \\
 \\
${}^1$ Extreme Robotics Laboratory, Université de Birmingham, Edgbaston, B15 2TT, UK.   \\
${}^2$ Sorbonne Universit\'e, CNRS UMR 7222, INSERM U1150, ISIR, F-75005, Paris, France.
{} \\
 \\
m.k.j.adjigble@bham.ac.uk\\
}
\maketitle
\thispagestyle{empty}
\subsection*{Résumé}
{\em
Cet article présente un schéma d'asservissement visuel 3D direct pour l'alignement automatique de nuages de points (respectivement, objets) en utilisant des informations visuelles dans le domaine spectral. Plus spécifiquement, nous proposons une méthode d'alignement pour modèles/nuages de points 3D fonctionnant en estimant la transformation globale entre un nuage de point de référence et un nuage de point cible à l'aide d'une analyse des données dans le domaine harmonique. Une transformation de Fourier discrète 3D (TFD) dans $\mathbb{R}^3$ est utilisée pour l'estimation de la translation et les harmoniques sphériques réelles dans $\bm{SO(3)}$ sont utilisées pour l'estimation de la rotation. Cette approche nous permet de dériver un contrôleur par asservissement visuel découplé à 6 degrés de liberté. Nous montrons ensuite comment cette approche peut être utilisée comme contrôleur d'un bras robotique pour exécuter une tâche de positionnement. Contrairement aux méthodes existantes d'asservissement visuel 3D, notre méthode fonctionne bien avec des nuages de points partiels et dans le cas de grandes transformations initiales entre la position initiale et désirée. De plus, l'utilisation de données spectrales (au lieu de données spatiales) pour l'estimation de la transformation rend notre méthode robuste au bruit induit par les capteurs et aux occlusions partielles. Notre méthode a été validée expérimentalement avec succès sur des nuages de points obtenus à l'aide d'une caméra de profondeur montée sur un bras robotique.  
}
\subsection*{Mots-Clés}
Nuages de points, asservissement visuel, calcul de pose, recalage 3D, Transformée de Fourier 

 \subsection*{Abstract}
 {\em
 This paper presents a direct 3D visual servo scheme for the automatic alignment of point clouds (respectively, objects) using visual information in the spectral domain. Specifically, we propose an alignment method for 3D models/point clouds that works by estimating the global transformation between a reference point cloud and a target point cloud using harmonic domain data analysis. A 3D discrete Fourier transform (DFT) in $\mathbb{R}^3$ is used for translation estimation and real spherical harmonics in $\bm{SO(3)}$ are used for rotation estimation. This approach allows us to derive a decoupled visual servo controller with 6 degrees of freedom. We then show how this approach can be used as a controller for a robotic arm to perform a positioning task. Unlike existing 3D visual servo methods, our method works well with partial point clouds and in cases of large initial transformations between the initial and desired position. Additionally, using spectral data (instead of spatial data) for the transformation estimation makes our method robust to sensor-induced noise and partial occlusions. Our method has been successfully validated experimentally on point clouds obtained with a depth camera mounted on a robotic arm.
 }
 \subsection*{Keywords}
 Point clouds, visual servoing, pose estimation, 3D registration, Fourier Transform 

\section{Introduction}\label{sec:intro}
%
Au cours des trois dernières décennies, l'accent a été mis de plus en plus sur les méthodes d'asservissement visuel pour réaliser des tâches robotiques dans divers secteurs comme l'industrie, la défense, les véhicules autonomes, la médecine, et bien d'autres. L'asservissement visuel désigne le contrôle dynamique de systèmes en utilisant un retour visuel continu. Par conséquent, les principales composantes d'un contrôleur d'asservissement visuel classique sont l'extraction des caractéristiques visuelles, leur mise en correspondance et leur suivi dans le temps. Cependant, la faisabilité et l'efficacité des méthodes d'asservissement visuel classiques sont étroitement liées à celles de détection, de mise en correspondance et de suivi d'informations visuelles, dont les performances peuvent être limitées dans certains cas (images peu texturées, absence de formes géométriques saillantes, occlusions, etc.).
\begin{figure}[!h]
    \centering
    \includegraphics[width=0.95\columnwidth]{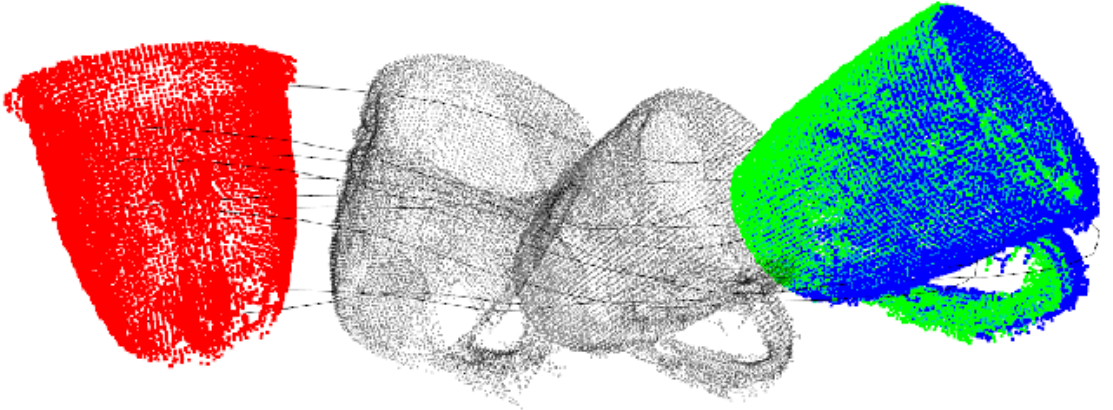}
    \caption{
    Illustration du processus d'alignement de modèle avec l'approche proposée. Le nuage de points rouge est le modèle de référence, les nuages de points gris sont des candidats intermédiaires au cours de la convergence, le nuage de point vert est le nuage de points de la scène cible et en bleu, le modèle final aligné sur la cible. Les courbes de trajectoire grises indiquent la convergence du modèle.}
    \label{fig:fig1}
\end{figure}
Des méthodes d'asservissement visuel qui s'affranchissent du suivi visuel ont récemment émergé. Elles utilisent directement l'information globale de l'image dans la boucle de commande. Ces approches sont appelées méthodes \emph{d'asservissement visuel direct}~\cite{deguchi2000direct}. Différents types d'informations globales ont été étudiées dans la littérature, telles que l'intensité des pixels~\cite{collewet2011photometric, Tamadazte2012}, les gradients spatio-temporels~\cite{Marchand2010gradient}, les histogrammes~\cite{bateux2016histograms}, l'information mutuelle~\cite{dame2011mutual}, les mélanges de Gaussiennes~\cite{crombez2018visual}, etc. Plus récemment, certains auteurs ont proposé d'utiliser des informations visuelles temps-fréquence en utilisant les ondelettes~\cite{ourak2019direct} et shearlets~\cite{duflot2019wavelet}. Cependant, les méthodes directes présentent clairement des domaines de convergence plus étroits par rapport aux méthodes conventionnelles. Pour remédier à ce problème, certains travaux proposent d'utiliser des caractéristiques visuelles exprimées dans le domaine spectral. Ces caractéristiques se sont avérées robustes au bruit et sont utilisées pour de nombreuses applications de vision par ordinateur et robotique comme la corrélation d'images~\cite{larsson2011correlating}, l'alignement de modèles issus de capteurs de profondeurs~\cite{bulow2012spectral}, la préhension robotique~\cite{adjigble2021spectgrasp}, etc. Dans~\cite{marchand2020direct}, les coefficients de la transformée en cosinus discrète (TCD) ont été utilisées pour l'asservissement visuel direct et dans~\cite{marturi2014visual, marturi2016image}, la propriété de translation temporelle de la transformée de Fourier a été utilisée dans 
 un schéma de commande découplé par asservissement visuel.

La majorité des méthodes d'asservissement visuel présentées ci-dessus sont fondées sur l'utilisation d'information visuelles 2D. Cependant, très peu sur l'asservissement visuel utilisant des informations 3D directement dans la boucle de commande~\cite{teuliere2014dense, dahroug2020pca}. Un des avantages à utiliser des informations visuelles 3D est notamment l'expression direct du contrôleur sans passer par des procédures d'estimation de pose ou d'étalonnage. Bien que ces méthodes aient donné des résultats prometteurs, elles nécessitent des données de profondeur denses et présentent une convergence limitée. Une méthode d'asservissement visuel virtuelle utilisant un maillage polygonale généré hors ligne à partir de nuages de points est présentée dans~\cite{kingkan2016model}.

Dans cet article, nous présentons une méthode d'asservissement visuel 3D direct utilisant des informations visuelles exprimées dans le domaine spectral. Bien que très peu de travaux aient utilisé l'information spectrale dans une loi de commande par asservissement visuel 2D~\cite{marchand2020direct, ourak2016wavelets, marturi2016image, marturi2014visual}, à notre connaissance il n'y a pas de travaux dans la littérature qui utilisent des informations spectrales 3D issues de nuages de points. L'idée principale de notre approche est l'estimation d'une transformation spatiale (6 DDLs) entre un nuage de point de référence et une autre cible. Les deux nuages de points transformés dans le domaine spectral, ainsi la translation est estimée par analyse de Fourier alors que la rotation est estimée par corrélation sphérique. Une procédure d'optimisation a été ensuite utilisée pour itérativement minimiser la fonction coût en translation et en rotation. A noter que l'estimation des translations et rotations de l'objet sont indépendantes l'une de l'autre. Cela nous permet d'avoir un contrôleur complètement découplé. La méthode proposée utilise une transformée de Fourier rapide 3D dans l'espace Cartésien $\mathbb{R}^3$ et les harmoniques sphériques réelles sur la sphère unitaire $\bm{S}^2$ et le groupe de rotation $\bm{SO(3)}$ pour respectivement calculer le gradient des coûts de translation et de rotation. 

Un exemple d'alignement d'un objet dans une scène simple est présenté à la Fig.~\ref{fig:fig1}, où un modèle de référence est aligné sur un nuage de points de scène contenant un seul et même objet. Le nuage de points de la scène actuelle ou de la scène cible est capturé en ligne par un capteur de profondeur statique ou monté ou sur un bras robotique. 

Les principales contributions de cet article sont les suivantes : 
\begin{itemize}
    \item Nous proposons une nouvelle méthode fondée sur le domaine spectral pour l'alignement des modèles d'objets en 3D, c'est-à-dire pour estimer la translation et rotation globales entre deux nuages de points.
    
    \item Nous proposons une nouvelle méthode d'asservissement visuel 3D  direct à 6 DDls utilisant directement des nuages de points (complets ou partiels) représentés dans le domaine spectral. 
\end{itemize}

Les avantages de la méthode proposée sont multiples. Contrairement aux approches existantes qui nécessitent des données de profondeur denses et complètes, notre méthode peut fonctionner efficacement avec des nuages de points partiels d'objets. En comparaison aux méthodes d'asservissement visuel directes utilisant des données de profondeur, notre méthode montre un domaine de convergence plus large. L'utilisation de données spectrales rend naturellement notre méthode robuste face au bruit, plus particulièrement lorsqu'on utilise des nuages de points issus de capteurs de distance 3D. Étant donné qu'aucune information de couleur ou d'intensité n'est requise, la méthode proposée peut bien fonctionner dans le cas d'objets non-texturés ou dans le cas d'éclairage faible. Finalement, la méthode proposée peut être utilisée à la fois pour aligner un modèle d'objet dans une scène avec plusieurs objets différents et pour positionner un manipulateur robotique dans l'espace de tâche.
%
\section{Méthodologie}\label{sec:method}
%
Dans cette section, nous présentons notre méthode d'asservissement visuel direct 3D dans le domaine spectral. Comme mentionné précédemment, la base principale de notre approche est la stratégie d'alignement de modèles d'objets à l'aide de nuages de points transformés dans le domaine spectral. Dans cette optique, nous introduisons tout d'abord la représentation utilisée par notre méthode, ensuite les concepts de corrélation de phase dans l'espace Cartésien et sur la sphère unitaire. Enfin, la loi de contrôle d'alignement itératif de modèle est présentée.
%
\subsection{Représentation de Nuage de Points}
La première étape de notre pipeline d'asservissement visuel 3D consiste à représenter les points et normales de surface du nuage de points respectivement comme une grille de voxels et une Image Gaussienne Étendue (IGE ou Extended Gaussian Image).

\subsubsection{Points sous forme de grille de voxels}

La discrétisation d'un nuage de points est un processus simple. Étant donné un nuage de points composé de $N$ points, une grille de voxels 3D de résolution $r \in \mathbb{R}^+$ peut être construite. Ainsi, pour chaque point $p = (x, y, z)$ du nuage de points, les indices du voxel du point $p_{ijk}= (i, j, k)$ sont calculés comme suit : 
\begin{equation}
\label{eq:voxel_index}
i = [x/r]\qquad
    j = [y/r]\qquad
    k = [z/r]
\end{equation}

L'opération $[./.]$ représente la division d'entier, c'est-à-dire que seule la partie en entière de la division est conservée.
Définissons $v_t: \mathbb{R}^3 \rightarrow \mathbb{N}^3$ comme étant la fonction de correspondance entre les coordonnées Cartésiennes et les indices de voxels. La fonction de grille de voxel\footnote{L'indice $t$ indique que la fonction est utilisée pour l'estimation de la translation, de la même manière, l'indice $r$ sera utilisé pour les fonctions liées à l'estimation de la rotation.} $f_t: \mathbb{R}^3 \rightarrow \mathbb{N}$ d'un nuage de points peut être défini comme suit : 
\begin{equation}
f_t(p) = f_t(x, y, z) = v_{ijk}
\label{eq:voxel}
\end{equation}
où $v_{i,j,k} \in [0,1]$. Ici, $v_{i,j,k}=1$ si au moins un point du nuage de points a des indices égaux à $v_T(p) = (i,j,k)$ et $v_{ijk} = 0$ dans le cas contraire. Notre méthode utilise une grille de voxels à valeurs réelles ; cependant, une grille de voxels à valeurs binaires peut aussi être utilisée de la même manière. Le score LoCoMo (Local Contact Moments) présenté dans~\cite{adjigble2018model} peut être un bon candidat pour améliorer l'information contenue dans les voxels.
\subsubsection{Normales de surface comme Image Gaussienne Étendue (IGE)}
l'IGE est une représentation populaire et utilisée pour les fonctions exprimées sur la sphère unitaire. Elle a été très utilisée dans la littérature comme descripteur de forme pour les normales de surface des objets \cite{little1985extended,nayar1990specular,lowekamp2002exploring,adjigble2021spectgrasp}. Changer la représentation d'une normale de surface $n = (n_x, n_y, n_z) \in \mathbb{R}^3$ utilisant des coordonnées Cartésiennes aux coordonnées sphériques $n = (r, \theta, \phi)$ en utilisant \eqref{eq:spherical_coordinates}, permet d'exprimer la normale de surface sur la sphère unitaire.
\begin{equation}
\label{eq:spherical_coordinates}
\begin{gathered}
    r      = \sqrt{n_x^2 + n_y^2 + n_z^2} \qquad
    \theta = \arctan{\frac{\sqrt{n_x^2 + n_y^2}}{n_z}} \\
   \phi    = \arctan(\frac{n_y}{n_x}) 
\end{gathered}
\end{equation}

\begin{figure}[!h]
    \centering
    \includegraphics[width=0.9\columnwidth]{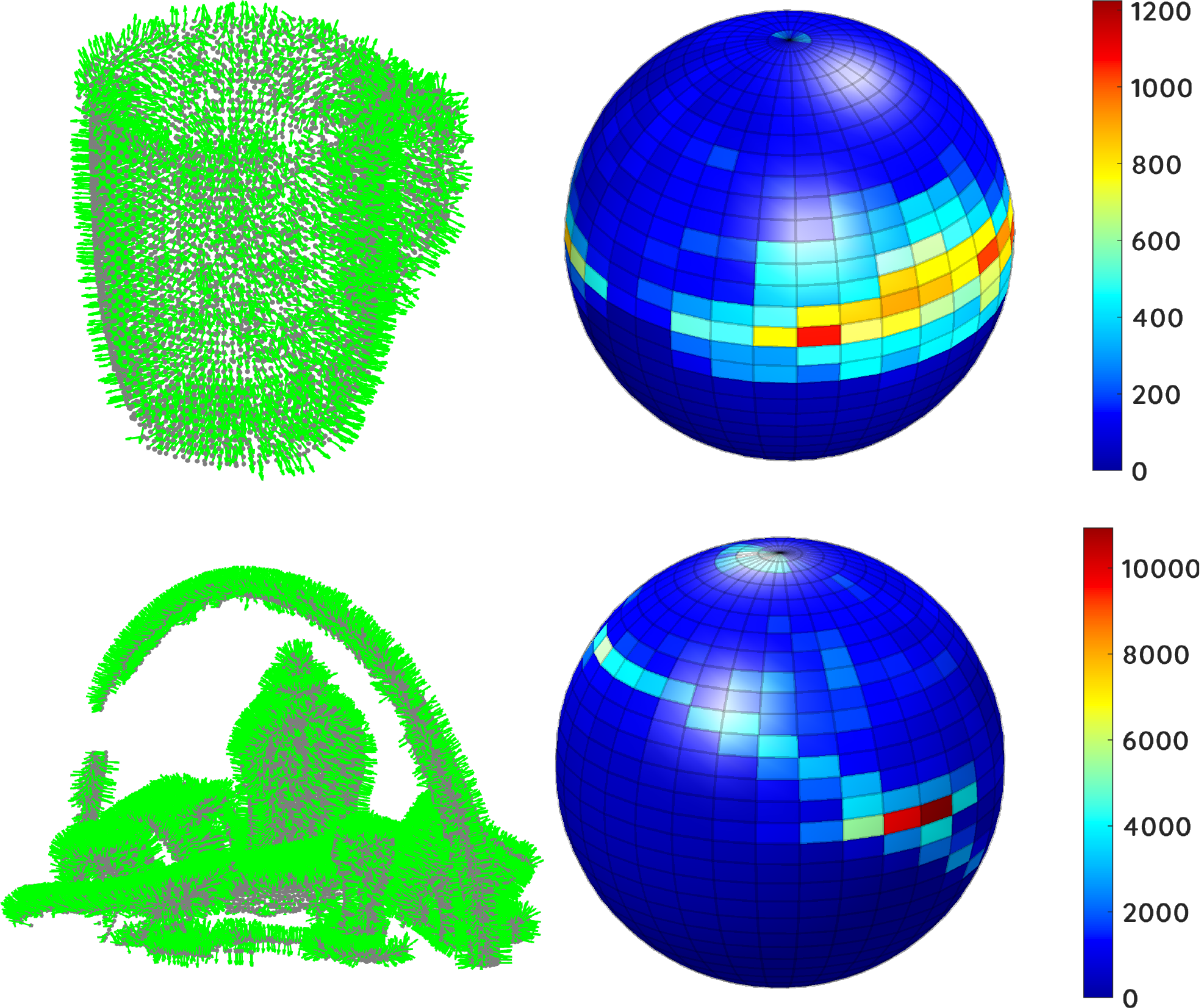}
    \caption{IGE d'une tasse (haut) et d'une scène à plusieurs objets (bas), qui sont représentées sous forme de nuages de points avec des normales de surface (petites flèches vertes).}
    \label{fig:fig2_egi}
\end{figure}

La distance radiale $r = 1$ pour toutes les normales de surface puisqu'elles sont des vecteurs unitaires. De ce fait, l'ensemble $(\theta, \phi)$ suffit pour décrire la distribution des normales de surface sur la sphère unitaire.
Une représentation discrète de la sphère est nécessaire pour effectuer les calculs numériques. La discrétisation suivante est utilisée en fonction de la longitude et latitude : $\theta_j = \frac{\pi(2j + 1)}{4B}$ et $\phi_k = \frac{\pi k}{B}$, $(j, k) \in \mathbb{N}$ avec la contrainte $0  \leq j, k < 2B$ et $B \in \mathbb{N}$, $B$ étant la bande passante. La valeur de la bande passante est habituellement choisie comme puissance de $2$, ce qui signifie que $B=2^n, n \in \mathbb{N}^+$. L'IGE des normales de surface d'un nuage de points peut ainsi être exprimée comme la fonction $f_r : \bm{S}^2 \rightarrow \mathbb{N}$ :
\begin{equation}
f_r(\theta, \phi) = c_r({\theta_j, \phi_k})
\label{eq:egi}
\end{equation}
où $c_r \in \mathbb{N}$ représente le nombre de normales de surface dans le nuage de points avec une longitude et une latitude discrétisée égales à $(\theta_j, \phi_k)$. Dans ce cas, des valeurs entières sont utilisées plutôt que des valeurs binaires. L'avantage est qu'une distribution des normales de surface sur la sphère unitaire fournit plus d'information sur la géométrie de l'objet qu'une simple distribution binaire. Figure~\ref{fig:fig2_egi} montre des exemples d'IGE d'un objet et d'une scène avec plusieurs objets, représentés sous forme de nuages de points.
%

\subsection{Estimation de Translation via l'Analyse de Fourier dans $\mathbb{R}^3$}
%
La translation entre deux nuages de points (cible et référence) est estimée à l'aide de la corrélation de phase 3D dans le domaine spectral avec l'analyse de Fourier. Le principal avantage des méthodes fondées sur l'analyse de Fourier est qu'elles sont robustes aux différents types de bruit~\cite{bulow2012spectral, marturi2016image}. La méthode de corrélation de phase est fondée sur la propriété de translation temporelle de Fourier et convertit les translations dans l'espace Cartésien en déplacement de phase dans le domaine spectral.

Définissons $f_t : \mathbb{R}^3 \rightarrow \mathbb{N}$ comme étant la représentation en voxels du nuage de points d'un objet ou d'une scène.
Les coefficients de Fourier de $f_t$ sont calculés comme suit : 
\begin{equation}
F_t(u, v, w) = \sum_{x=0}^{M-1}\sum_{y=0}^{N-1}\sum_{z=0}^{L-1}f_t(x,y,z)e^{-i2\pi(\frac{u}{M}x + \frac{v}{N}y + \frac{w}{L}z)}
\label{eq:fourier_coefs}
\end{equation}
où, $M, N, L \in N^+$ sont les degrés maximaux de décomposition des coefficients de Fourier sur respectivement sur $X$, $Y$, et $Z$ et $(u, v, w)$ sont les coordonnées correspondantes dans le domaine spectral. Supposons que l'objet ou la scène soit translatée par $T = (\tau_x,\tau_y, \tau_z) \in \mathbb{R}^3$, et que $g_t : \mathbb{R}^3 \rightarrow \mathbb{N}$ est la nouvelle représentation en voxels du nuage de points translaté. En s'appuyant sur la propriété de translation temporelle, les coefficients de Fourier $G_t$ de $g_t$ peuvent être calculés à l'aide de : 
\begin{equation}
G_t(u, v, w) = F_t(u, v, w)e^{-i2\pi(\frac{u}{M}\tau_x + \frac{v}{N}\tau_y + \frac{w}{L}\tau_z)}
\label{eq:fourier_shift_theorem}
\end{equation}

Le but de l'estimation de la translation est de trouver $T$ en connaissant $f_t$ et $g_t$. Ce qui peut être réalisé objectivement en tout d'abord calculant le spectre de puissance croisé normalisé $\mathcal{C}_t$ de $F_t$ et $G_t$ et en appliquant la transformée de Fourier inverse en utilisant \eqref{eq:cross_power_spectrum}.
\begin{equation}
\begin{aligned}
\mathcal{C}_t(u,v,w) &= \frac{F_t(u,v,w)\overline{G_t(u,v,w)}}{|F_t(u,v,w)\overline{G_t(u,v,w)}|} \\
\delta(\tau_x,\tau_y, \tau_z) &= \mathcal{F}^{-1}(\mathcal{C}_t(u,v,w)) 
\label{eq:cross_power_spectrum}
\end{aligned}
\end{equation}
où, $\overline{G_t}$ est le complexe conjugué de $G_t$ et $\mathcal{F}^{-1}$ est la transformée de Fourier inverse. Le résultat $\delta$ est la fonction \emph{delta de Dirac} dont la position sur l'axe des abscisses de la valeur maximale correspond à la translation $T$. Par conséquent, la translation $T$ peut être trouvée en maximisant la fonction $\delta$.
\begin{equation}
\begin{aligned}
T &= \nabla_{glob} T = \mathrm{argmax}\{\delta(\tau_x,\tau_y, \tau_z)\}
\label{eq:fourier_shift_solution}
\end{aligned}
\end{equation}

Même si la solution globale $\nabla_{glob} T$ de la translation peut être trouvée directement, dans le contexte d'asservissement visuel 3D, seule une petite partie du vecteur égale à $\nabla T = \lambda_t\nabla_{glob} T$, avec $\lambda_t \in \mathbb{R}^+$ et $\lambda_t < 1$, sera utilisée à chaque itération. Cela permet d'estimer simultanément la translation et la rotation, mais aussi de contrôler le taux de convergence du contrôleur.
La fonction de coût $J_t(T)$ suivante peut être formulée pour évaluer la performance de l'algorithme d'estimation de la translation sur $\mathbb{R}^3$ : 
\begin{equation}
\begin{aligned}
J_t(T) = \frac{1}{2}||g_t(x) - f_t(x + T)||^2 \\
\end{aligned}
\label{eq:translation_cost_function}
\end{equation}
%
\subsection{Estimation de la Rotation par Analyse de Fourier dans $\bm{S}^2$}
Comme pour la translation, la rotation entre deux nuages de points (cible et référence) peut aussi être estimée en utilisant une analyse spectrale.
Ici, la représentation unitaire de données exprimées sur la sphère unitaire est utilisée pour encoder l'information des normales de surface de l'objet. Dans ce cas, nous estimons la rotation globale à partir de la corrélation d'IGE. Il est possible de trouver la rotation "exacte" directement en cherchant la valeur de la rotation qui maximise la corrélation, mais cela implique le calcul d'une intégrale double gourmande en temps de calcul. Comme solution, une optimisation de type gradient est utilisée pour calculer de manière itérative la rotation qui maximise la corrélation entre les deux nuages de points.
\subsubsection{Transformée de Fourier dans $\bm{S}^2$ et $\bm{SO}(3)$}
Soit $f_r : \bm{S}^2 \rightarrow \mathbb{N}$ l'IGE des normales de surface d'un objet. De fait que les valeurs de la fonction $f_r$ sont dans l'ensemble $\mathbb{N} \subset \mathbb{R}$, l'analyse harmonique réelle dans $\bm{SO(3)}$, introduite dans~\cite{lee2018real}, peut être utilisée pour calculer les coefficients de Fourier. Étant donnée une bande passante $B$, la transformée de Fourier de $f_r$ dans $\bm{S}^2$ est définie par : 
\begin{equation}
f_r(\theta, \phi) = \sum_{l=0}^{B-1}(F^l_r)^TS^l(\theta, \phi)
\label{eq:real_fourier_transform_fr}
\end{equation}
où, $F^l_r \in \mathbb{R}^{(2l+1)\times1}$ sont les coefficients de la transformée de Fourier et $S^l \in \mathbb{R}^{2l+1}$ sont les axes orthogonaux de fonctions à valeurs réelles définies sur $\bm{S}^2$. Le vecteur $S^l$ est construit à partir des harmoniques sphériques réelles $Y^l(\theta, \phi$) et d'une matrice $T^l \in \mathbb{C}^{(2l+1)\times(2l+1)}$ à coefficients complexes comme suit : 
\begin{equation}
S^l(\theta, \phi) = T^lY^l(\theta, \phi)
\label{eq:real_fourier_transform}
\end{equation}
Se référer à~\cite{lee2018real, blanco1997evaluation} pour plus de détails sur les harmoniques sphériques.

Supposons que le nuage de points soit transformé par une rotation $R \in SO(3)$ autour de son centre de gravité. $R$ est paramétrée en utilisant la convention $ZYZ$ des angles d'Euler par $\alpha, \gamma \in [0,2\pi[$ et $\beta \in [0, \pi]$, avec $g_r : \bm{S}^2 \rightarrow \mathbb{N}$ étant l'IGE du nuage de point transformé par la rotation. La matrice de rotation $R$ peut ainsi être exprimée comme :
\begin{equation}
R = R(\alpha, \beta, \gamma) = \exp(\alpha\hat{e}_z)\exp(\beta\hat{e}_y)\exp(\gamma\hat{e}_z)
\label{eq:euler_representation}
\end{equation}
avec $e_y$ et $e_z$ étant respectivement les vecteurs $(0,1,0)$ et $(0,0,1)$. L'opérateur $\hat{.}: \mathbb{R}^3 \rightarrow \mathfrak{so}(3)$ transforme un vecteur 3D en sa matrice symétrique de dimension $3\times3$ par l'algèbre de Lie: $\mathfrak{so}(3) = \{S \in R^{3\times3} | S + S^T = 0\}$. Bien que la représentation de \eqref{eq:euler_representation} présente des singularités inhérentes, elle est très pratique pour calculer de la transformée de Fourier dans $\bm{SO(3)}$. De même que dans \eqref{eq:real_fourier_transform_fr}, la transformée de Fourier de $g_r$ est obtenue par :
\begin{equation}
g_r(\theta, \phi) = \sum_{l=0}^{B-1}(G^l_r)^TS^l(\theta, \phi) 
\label{eq:real_fourier_transform_gr}
\end{equation}
avec $G^l_r \in \mathbb{R}^{(2l+1)\times1}$ sont les coefficients de Fourier.

En considérant que $g_r$ est une version transformée de $f_r$ par une rotation, d'où la relation~\eqref{eq:gr}, la transformée de Fourier de $g_r$ peut être calculée en utilisant les coefficients de Fourier de $f_r$ par~\eqref{eq:real_fourier_transform_gr_simplified}.
\begin{equation}
g_r(\theta, \phi) = f_r(R^T(\theta, \phi))
\label{eq:gr}
\end{equation}
où $R^T(\theta, \phi)$ est une notation simplifiée pour l'expression $M_{s2c}^{-1}(R^TM_{s2c}(\theta, \phi))$, où $M_{s2c}:\bm{S}^2 \rightarrow \mathbb{R}^3$ est la fonction qui convertit les coordonnées sphériques en coordonnées Cartésiennes, et $M_{s2c}^{-1}:\mathbb{R}^3 \rightarrow \bm{S}^2$, son inverse qui peut être obtenue par~\eqref{eq:spherical_coordinates}. 
En remplaçant $f_r$ par sa valeur et en reformulant \eqref{eq:gr}, on obtient :
\begin{equation}
\begin{aligned}
g_r(\theta, \phi) &= \sum_{l=0}^{B-1}(F^l_r)^TS^l(R^T(\theta, \phi)) \\
&= \sum_{l=0}^{B-1}(U^l(R)F^l_r)^TS^l(\theta, \phi)
\end{aligned}
\label{eq:real_fourier_transform_gr_simplified}
\end{equation}
où, $U^l(R)=\overline{T^l}D^l(R)(T^l)^T$. $\overline{T^l}$ est le conjugué complexe de $T^l$, tandis que $\bm{D}^l$ est la matrice $\bm{D}$ de Wigner. L'expansion de \eqref{eq:real_fourier_transform_gr_simplified} est possible puisque les rotations sont exprimées comme des matrices $\bm{D}$ de Wigner dans le domaine spectral et appliquer une rotation aux fonctions de base $S^l$ est équivalente à appliquer une transformation linéaire des fonctions de base par la matrice $\bm{D}$ de Wigner associée à la rotation. De \eqref{eq:real_fourier_transform_gr} et \eqref{eq:real_fourier_transform_gr_simplified}, nous constatons que $G^l_r = U^l(R)F^l_r$. Ainsi, $G^l_r$ est obtenu en appliquant la transformation $U^l(R)$ aux coefficients de Fourier de $f_r$. Pour plus de détails sur les propriétés couramment utilisées de la matrice $\bm{D}$ de Winger, se référer à \cite{lee2018real, blanco1997evaluation, kostelec2008ffts}. L'objectif de l'estimation de rotation est de trouver $R$ en connaissant $f_r$ et $g_r$.
\subsubsection{Corrélation dans $\bm{SO}(3)$ et ses dérivées}
La corrélation entre $f_r$ et $g_r$ est calculée comme suit : 
\begin{equation}
\mathcal{C}_r(R) = corr(f_r,g_r) = \frac{1}{4\pi}\sum_{l=0}^{B-1}(G^l_r)^TU^l(R)F^l_r
\label{eq:correlation_rotation}
\end{equation}

Ce résultat est obtenu après simplification, en remplaçant $f_r$ et $g_r$ par leurs représentations de Fourier~\eqref{eq:real_fourier_transform_fr} et~\eqref{eq:real_fourier_transform_gr_simplified}, en utilisant le théorème de convolution de la transformée de Fourier\footnote{La convolution dans le domaine spatial équivaut à la multiplication des coefficients de Fourier dans le domaine spectral} et le principe d'orthogonalité des bases $S^l$. La relation $\langle S^l(\theta, \phi), (S^l(R^T(\theta, \phi))^T)\rangle=\frac{1}{4\pi}U^l(R)$ résulte directement de l'orthogonalité des vecteurs de base $S^l$, où l'opérateur $\langle . \rangle$ est le produit interne dans $\mathcal{L}^2(SO(3))$. Dans~\eqref{eq:correlation_rotation}, uniquement $U^l$ dépend de la rotation $R$, ainsi, la dérivée de $\mathcal{C}_r$ peut être obtenue en calculant la dérivée de $U^l$. La dérivée de $U^l$, évaluée à $R$, par rapport à une rotation élémentaire $R_{\epsilon} = \exp(\epsilon \hat{\eta})$ ($\epsilon \approx 0$ et $\eta \in \mathbb{R}^3$) est calculée comme suit : 
\begin{equation}
\left.\frac{d}{d \epsilon}\right|_{\epsilon=0} U^l(R \exp (\epsilon \hat{\eta}))= U^l(R)\left.\frac{d}{d \epsilon}\right|_{\epsilon=0} U^l(\exp (\epsilon \hat{\eta}))
\label{eq:ul_derivative}
\end{equation}

Dans l'équation précédente, la propriété d'homomorphisme de $U^l$ a été utilisée, c.-à-d., $U^l(R_1R_2) = U^l(R_1)U^l(R_2)$ pour $R_1, R_2 \in \bm{SO}(3)$. La dérivée de $\mathcal{C}_r$ est ensuite calculée par :
\begin{equation}
\begin{aligned}
\left.\frac{d}{d \epsilon}\right|_{\epsilon=0} \mathcal{C}_r(\exp(\epsilon\hat{\eta})) &= \frac{1}{4\pi}\sum_{l=0}^{B-1}(G^l_r)^TU^l(R)u^l(\eta)F^l_r \cdot \eta \\
&=\nabla \mathcal{C}_r(R, \eta) \cdot \eta
\end{aligned}
\label{eq:pre_cr_derivative}
\end{equation}
où, $\nabla \mathcal{C}_r(R, \eta) \in \mathbb{R}^3$ est le gradient de $\mathcal{C}_r(R)$ autour de l'axe $\eta$ et $u^l(\eta) = \left.\frac{d}{d \epsilon}\right|_{\epsilon=0} U^l(\exp (\epsilon \hat{\eta}))$. Évaluer le gradient $\nabla \mathcal{C}_r(R, \eta)$ à $\eta = e_x, e_y, e_z$ permet de trouver la rotation élémentaire, qui compose avec $R$, accroît la valeur de la corrélation $\mathcal{C}_r$. Plus formellement :
\begin{equation}
\left.\nabla \mathcal{C}_r(R, e_k)\right|_{k\in \{x, y, z\}} = \frac{1}{4\pi}\sum_{l=0}^{B-1}(G^l_r)^TU^l(R)u^l(e_k)F^l_r
\label{eq:cr_derivative}
\end{equation}

Le calcul de $u^l(e_k)$ est trivial puisqu'il s'agit d'une différenciation directe des entrées de la matrice $\bm{D}$ de Wigner pour laquelle une dérivée analytique est disponible dans~\cite{lee2018real}.

Une méthode de descente de gradient peut maintenant être utilisée pour itérativement rechercher la rotation idéale. Une fonction coût $J_r$ peut être formulée pour évaluer les performances de l'algorithme d'estimation de rotation dans $\bm{SO}(3)$ :
\begin{equation}
J_r(R) = \frac{1}{2}||g_r(\theta, \phi) - f_r(R^T(\theta, \phi))||^2
\label{eq:rotation_cost_function}
\end{equation}

\subsection{Loi de Commande}
Pour estimer la transformation $H = (R, T) \in \bm{SO}(3)\times\mathbb{R}^3$ entre les nuages de points courant et de référence, la loi de commande donnée par~\eqref{eq:control_law} est utilisée.
\begin{equation}
\begin{aligned}
T &= T + \lambda_t\nabla_{glob} T \\
R &= R\exp{(\lambda_r \widehat{\nabla \mathcal{C}_r})}
\end{aligned}
\label{eq:control_law}
\end{equation}
où, $\lambda_t, \lambda_r \in \mathbb{R}^+$ et $\lambda_t, \lambda_r <1$. $\nabla_{glob} T$ et $\nabla \mathcal{C}_r$ sont respectivement calculés à partir de~\eqref{eq:fourier_shift_solution} et \eqref{eq:cr_derivative}. A la première itération, matrice de rotation $R$ et et le vecteur de $T$ peuvent être initialisés aléatoirement ou fixés respectivement en une matrice identité et de zéros, respectivement. Le contrôleur converge lorsque :
\begin{equation}
    ||\nabla_{glob} T|| + ||\nabla \mathcal{C}_r|| < \epsilon_g
    \label{eq:updaterule}
\end{equation}
où, $\epsilon_g \in \mathbb{R}^+$ est le seuil de tolérance. Pour contrôler le robot, la loi de commande suivante est utilisée :
\begin{equation}
    \dot {\mathbf{q}} = \mathcal{\bm{J}}_c^+\dot{X_c}
    \label{eq:robot_controller}
\end{equation}
avec, $\mathcal{\bm{J}}_c^+$ étant le pseudo-inverse de la Jacobienne du robot exprimée dans le repère de la caméra, $\dot{{\mathbf{q}}}$ est le vecteur des vitesses articulaires du robot et $\dot{X_c}$ les vitesses Cartésiennes de la caméra, calculées à partir de \eqref{eq:control_law}. L'algorithme correspondant à la loi de commande est présenté dans Alg.~\ref{alg:algorithm}.
\begin{algorithm}[t]
\label{alg:algorithm}
 \caption{Asservissement visuel 3D direct dans le domaine spectral}
  Initialiser $R$ par une matrice identité\\
  Initialiser $T$ par un vecteur de zéros\\
  Initialiser les gains $\lambda_t, \lambda_r$ et le seuil
  $\epsilon_g$ \\
  Calculer $f_t$ \eqref{eq:voxel}, $f_r$ \eqref{eq:egi}, $F_t$ \eqref{eq:fourier_coefs}, $F^l_r$ du nuage cible \\
  \While{$||\nabla_{glob} T|| + ||\nabla \mathcal{C}_r|| >= \epsilon_g$}{
  Acquérir le nuage de points de la scène \\
  Calculer $g_t$ \eqref{eq:voxel}, $g_r$ \eqref{eq:egi}, $G_t$ \eqref{eq:fourier_coefs}, $G^l_r$ du nuage de points de référence \\
  Calculer $\nabla_{glob} T$ \eqref{eq:fourier_shift_solution} et $\nabla \mathcal{C}_r$ \eqref{eq:cr_derivative} \\
  Appliquer la règle de mise à jour \eqref{eq:control_law} \\
  Calculer fonction coût coût $J = J_t(T) + J_r(R)$ \\
  Contrôler le robot en utilisant \eqref{eq:robot_controller}\\
  }
  Récupérer la transformation finale $H = (R, T)$ \\

\end{algorithm}
%
\section{Validation Expérimentale} \label{sec:results}
%
\subsection{Description de la Plate-forme Expérimentale}
Les validations expérimentales sont réalisées à l'aide de nuages de points acquis grâce à une caméra de profondeur. Deux essais différents sont présentés dans cet article. Nous validons d'abord le processus de recalage de modèles, où un modèle de référence complet d'un objet est aligné sur un nuage de points de scène. Ensuite, nous montrons des essais d'asservissement visuel direct réalisés à l'aide d'un robot à 7 DDLs (KUKA iiwa) muni d'une caméra de profondeur (Ensenso N35) montée sur l'effecteur du robot. Dans ce cas, la totalité du nuage de points est utilisé pour positionner le robot à une position objet-cible. La librairie de nuages de points (PCL)~\cite{Rusu_ICRA2011_PCL} est utilisée pour le traitement des nuages de points, les librairies FFTSO3~\cite{lee2018real} et FFTW~\cite{frigo1998fftw} sont utilisées pour l'analyse spectrale.

Comme mentionné dans la Sec.~\ref{sec:intro}, le nuage de points de l'objet de référence, utilisé pour les expériences de recalage de modèles, est construit hors ligne en combinant plusieurs nuages de points issus de différent points de vues comme présenté dans~\cite{marturi2019dynamic}. En utilisant le capteur Ensenso N35, les normales de surface pour chaque point sont obtenues directement au moment de l'acquisition. Les nuages de points sont voxélisés avec une grille de résolution $8~\mathrm{mm}$, alors que les normales de surface sont discrétisées sur la sphère unitaire en utilisant une bande passante $B=16$. Le degré maximal d'expansion des coefficients harmoniques sur la sphère est $l_{max} = 32$. Ces valeurs sont estimées empiriquement et donnent de bons résultats en termes de vitesse de calculs et de précision de recalage. Les trois principaux facteurs qui agissent sur la vitesse de convergence de notre approche sont la résolution de la grille de voxels, la bande passante de l'IGE et les paramètres $\lambda_t$ et $\lambda_r$. Des grilles plus fines nécessitent le calcul d'un nombre plus élevé de coefficients de Fourier, ce qui ralentit la vitesse de convergence de l'algorithme. Avec les paramètres mentionnés précédemment, la vitesse de traitement actuelle de notre approche est en moyenne de $8.7~\mathrm{ms/iteration}$.
\begin{figure}[!h]
    \centering
    \includegraphics[width=\columnwidth]{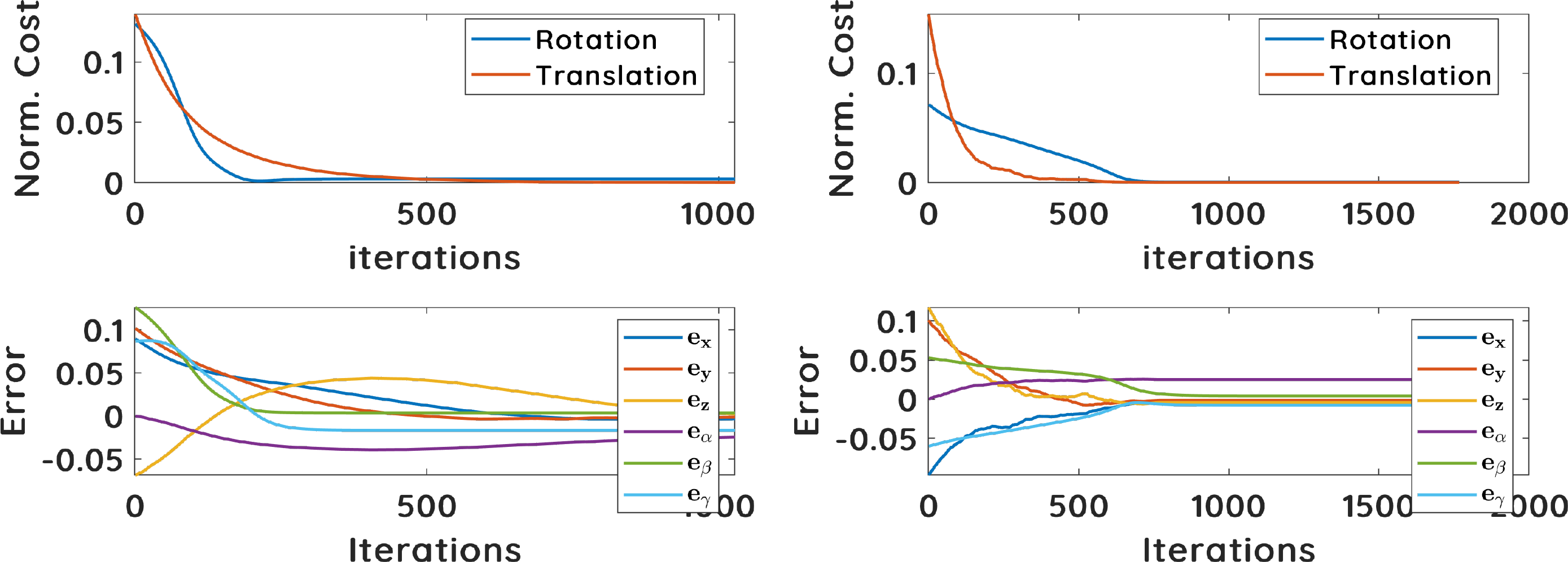}
    \caption{Courbes de convergence pour les objets: tasse (à gauche)  et  poignée de gaz (à droite) présentés dans Fig. \ref{fig:fig1}.}
    \label{fig:fig1_plots}
\end{figure}
%
%
\subsection{Analyse du recalage}
\begin{figure}[!h]
    \centering
    \includegraphics[width=\columnwidth]{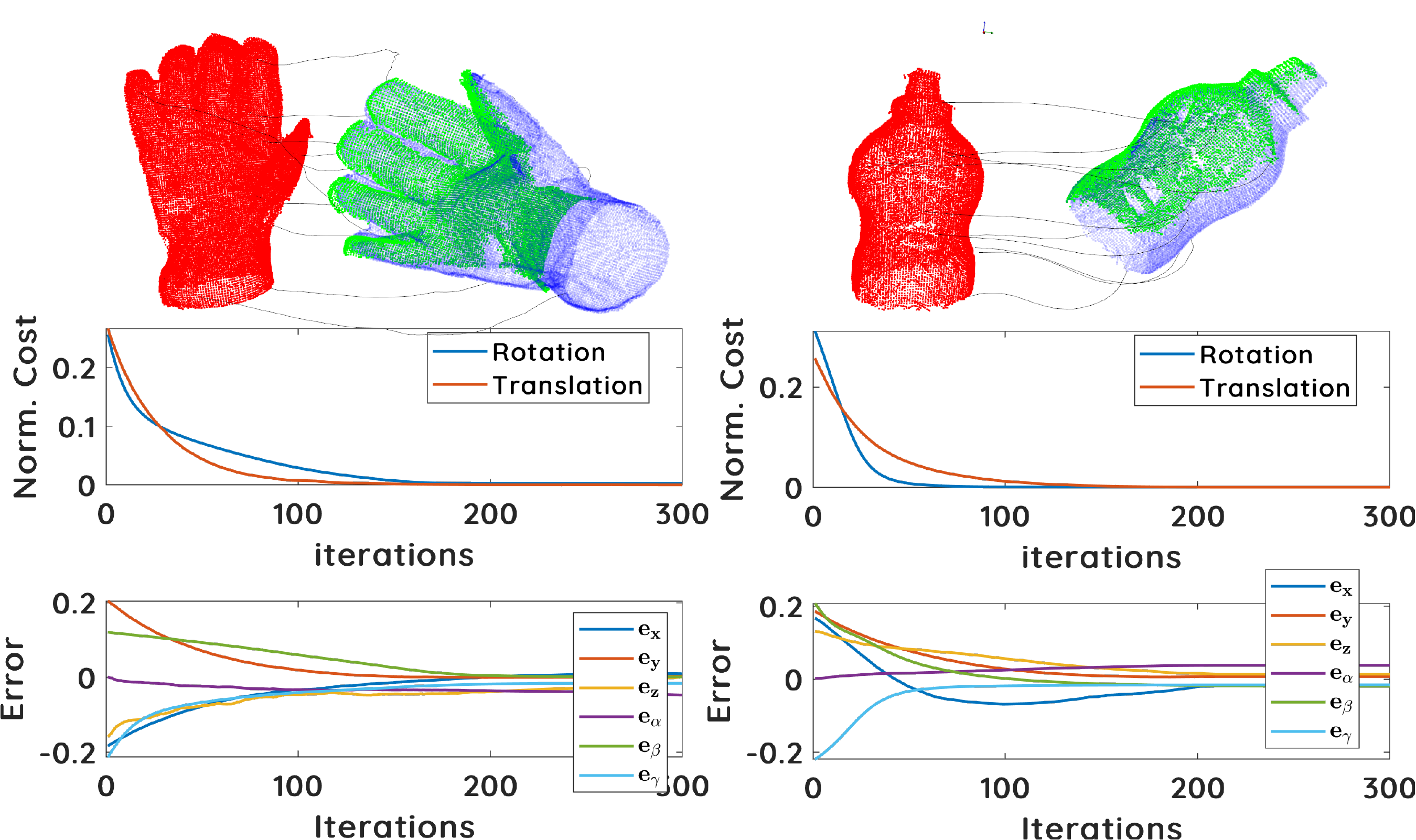}
    \caption{Exemple d'une procédure d'alignement dans le cas où un modèle de référence complet est aligné sur une scène partiellement observée. Les nuages de points rouges, bleus et verts représentent respectivement, la référence, le modèle aligné et la cible. Les résultats pour deux objets sont présentés pour un gant (à gauche) et une boîte de moutarde (à droite).}
    \label{fig:align_res}
\end{figure}

\begin{figure}[!h]
    \centering
    \includegraphics[width=\columnwidth]{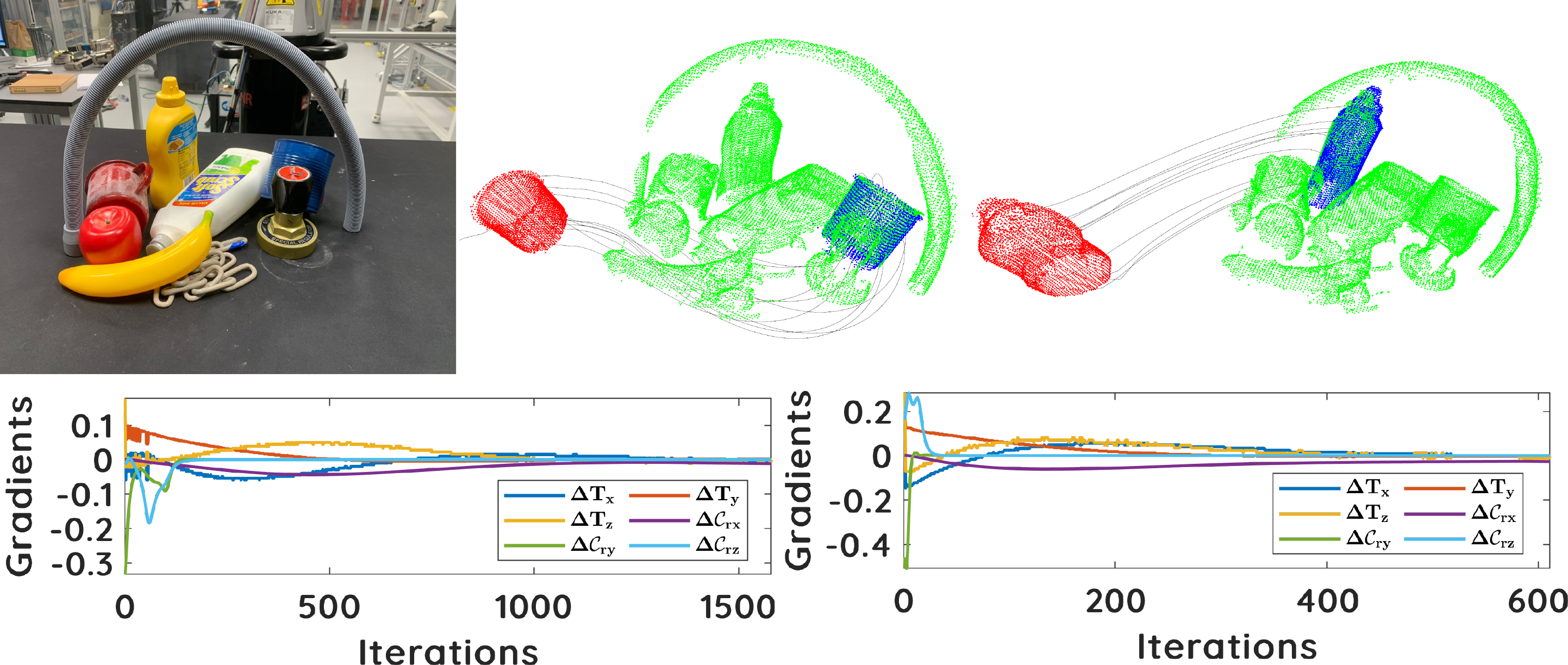}
    \caption{Exemple d'alignement de modèles dans le cas de scènes à plusieurs objets. La scène utilisée (en haut) pour l'alignement de deux objets différents ; et les courbes correspondantes (en bas) montrant l'évolution des gradients de translation et rotation pendant la procédure.}
    \label{fig:clutter_res}
\end{figure}

Les trois expériences suivantes sont réalisées pour valider la capacité de recalage de modèles de notre approche : (C–-1) le nuage de point complet d'un objet est aligné sur sa version transformée par une rotation et une  translation arbitraires (Fig.~\ref{fig:fig1} et Fig.~\ref{fig:fig1_plots}); (C-–2) le nuage de points complet d'un objet est recalé sur un nuage de point partiel cible du même objet (Fig.~\ref{fig:align_res}); et (C-–3) le nuage de points complet d'un objet est recalé sur une scène contenant plusieurs objets (Fig.~\ref{fig:clutter_res}).
Pour ces tests, les nuages de points de référence et cible partagent le même référentiel global. Différents objets de tous les jours sont utilisés pour les essais et les scènes avec plusieurs objets sont construites en positionnant plusieurs objets de manière aléatoire comme illustré sur la Fig.~\ref{fig:clutter_res}. Les images présentées dans les Fig.~\ref{fig:fig1}, \ref{fig:fig1_plots}, \ref{fig:align_res}, et \ref{fig:clutter_res}, montrent des exemples de convergence durant la procédure d'alignement et l'évolution des erreurs (translation et rotation) durant la procédure. 

A partir des résultats obtenus, on peut observer que les modèles de référence sont alignés avec précision avec les nuages de points cibles dans toutes les configurations. Les fonctions de coût de convergence finaux moyens pour les conditions C--1 et C--2 et la valeur moyenne finale du gradient pour C–3 sont présentés dans le tableau~\ref{tab:cost_table}. Les résultats démontrent clairement la précision de notre approche. En outre, notre méthode a démontré des performances tout à fait intéressantes dans des conditions complexes comme l'alignement d'un modèle complet à des modèles partiellement observés ainsi qu'à des scènes pas du tout structurées (avec des occlusions) contenant plusieurs objets. 

\midsepremove
\begin{table}[!h]
    \small
    \centering
    \caption{Valeurs moyennes à convergence.}
    \label{tab:cost_table}
    \begin{threeparttable}
    \begin{tabularx}{\columnwidth}{>{\hsize=1\hsize\raggedright\arraybackslash}X|
 >{\hsize=1\hsize\centering\arraybackslash}X|
 >{\hsize=1\hsize\centering\arraybackslash}X|
 >{\hsize=1\hsize\centering\arraybackslash}X}
         \toprule
         &  C--1 (coût) & C--2 (coût) & C--3 (gradient) \\
         \midrule
        Tran. erreur\tnote{1} & 1.775e-5 & 8.113e-5 & 6.4e-05  \\
        Rot. erreur\tnote{1} & 2.747e-2 & 3.2e-2& 1.199e-12  \\
        \bottomrule
    \end{tabularx}
    \begin{tablenotes}
        \item[1] \footnotesize{Calculé comme la moyenne de $(\mathrm{r\acute{e}elle} - 
 \mathrm{estimat\acute{e}e})^2$. \emph{Remarque}: ce calcul ne s'applique qu'à C--1 et C--2.}
    \end{tablenotes}
    \end{threeparttable}
\end{table}
\subsection{Tâche de Positionnement d'un Bras Robotique : Asservissement Visuel 3D Direct}
\begin{figure}[!h]
    \centering
    \includegraphics[width=\columnwidth]{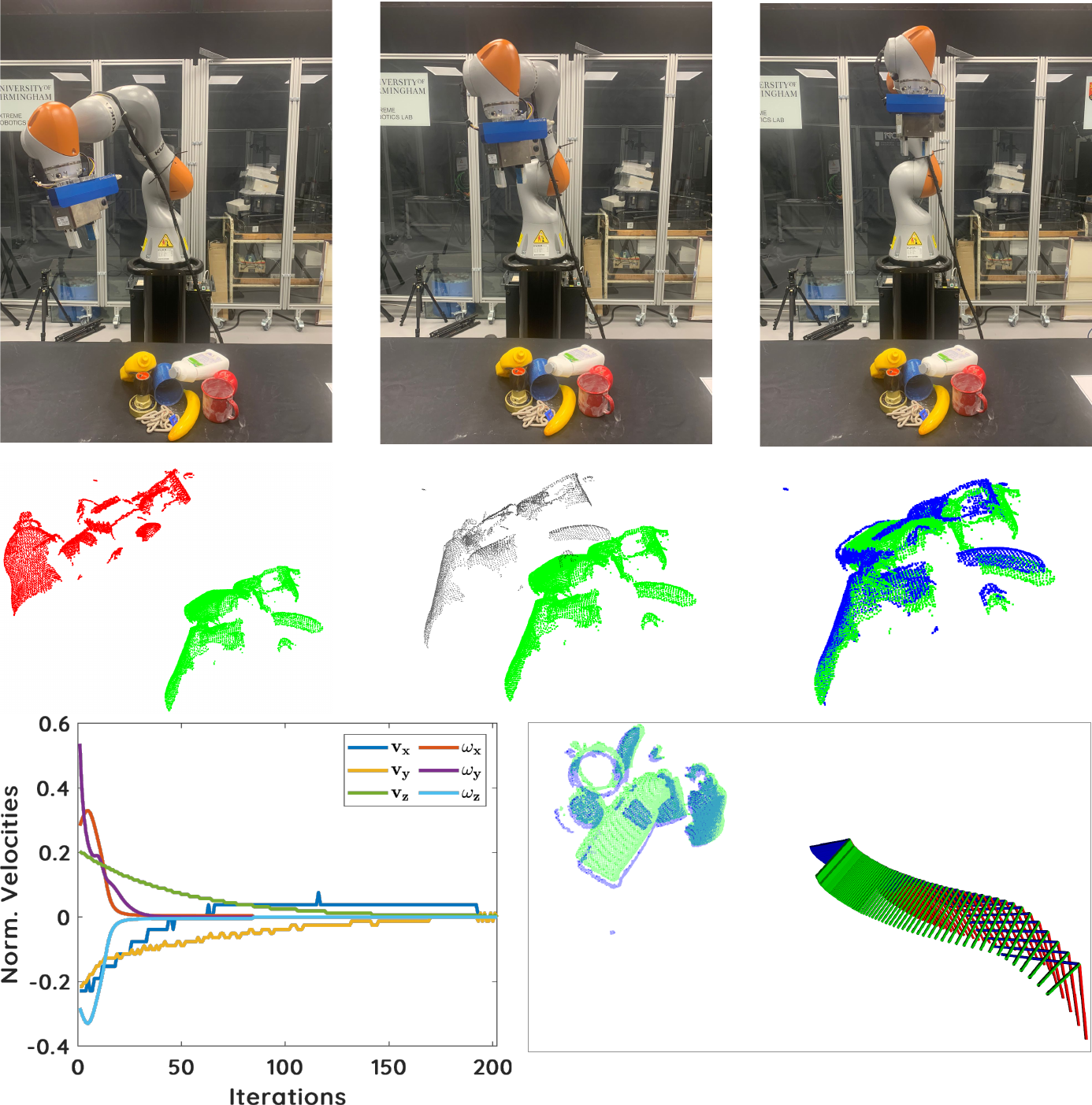}
    \caption{Illustration de l'asservissement visuel 3D direct. La rangée du haut montre le robot en position de départ, intermédiaire et finale. La rangée du milieu montre les nuages de points initial, intermédiaire et final. La rangée du bas montre la courbe de convergence et la trajectoire suivie par l'effecteur du robot. Pour ce test, le nuage de points complet est utilisé sans aucun appariement de modèle local.}
    \label{fig:dvs_result}
\end{figure}

Pour cet essai, nous considérons une tâche de positionnement pour laquelle la position du robot est contrôlée par notre méthode d'asservissement direct. Comme précédemment mentionné, l'ensemble du nuage de points est utilisé dans la boucle de commande, c'-à-d sans utiliser de méthode de segmentation. Ce test a été réalisé avec une scène constituée de plusieurs objets pour tester la généralité de notre approche, notamment dans le cas de scènes complexes. Le nuage de points de référence est initialement acquis à la position de désirée du robot. Le robot est ensuite déplacé à une position aléatoire dans l'espace des tâches en s'assurant d'une grande transformation spatiale. Les résultats obtenus sont illustrés sur la Fig.~\ref{fig:dvs_result}. A partir de ces résultats, il est clair que la méthode fonctionne bien lorsqu'il s'agit d'aligner un nuage de points entier à partir d'une position où seulement une partie de celle-ci est visible. Les courbes de convergence montrent la fluidité des mouvements du robot pour atteindre la position cible. 
%
\section{Conclusion} \label{sec:conclusion}

Dans cet article, nous avons présenté une méthode d'asservissement visuel 3D direct fondée sur l'utilisation de nuages de points exprimés dans le domaine spectral. L'approche présentée utilise la propriété de translation temporelle de la transformée de Fourier pour estimer les translations et les harmoniques sphériques réelles sur $\bm{SO(3)}$ pour estimer les rotations, afin d'aligner progressivement un nuage de points de référence sur un nuage de points cible. Cette approche a été initialement utilisée pour aligner des modèles 3D sur différentes scènes, puis utilisée pour contrôler la position d'un bras robotique pour réaliser automatiquement une tâche de positionnement. Les résultats expérimentaux obtenus démontrent l'efficacité de notre approche en termes de précision et de comportement du contrôleur y compris dans des conditions défavorables (exemple, occlusions).

Les travaux futurs seront axés sur l'utilisation de l'approche proposée pour la manipulation d'objets statiques ou en mouvement.

\end{document}